\documentclass[journal,twoside,web]{ieeecolor}
\usepackage{generic}
\usepackage{cite}
\usepackage{amsmath,amssymb,amsfonts}
\usepackage{algorithmic}
\usepackage{graphicx}
\usepackage{textcomp}

\usepackage{multirow}

\usepackage{hyperref}
\hypersetup{
    colorlinks=true,
    linkcolor=red,
    filecolor=magenta,      
    urlcolor=magenta,
    citecolor=blue,
}
\def\BibTeX{{\rm B\kern-.05em{\sc i\kern-.025em b}\kern-.08em
    T\kern-.1667em\lower.7ex\hbox{E}\kern-.125emX}}
\begin{document}
\title{SSR-HEF: Crowd Counting with Multi-Scale Semantic Refining and Hard Example Focusing}
\author{Jiwei Chen$^{*}$, Kewei Wang$^{*}$, Wen Su, Zengfu Wang
\thanks{$^{*}$ These authors have contributed equally to this work.}
\thanks{This work was supported by the Special Project of Strategic Leading Science and Technology of Chinese Academy of Sciences (No. XDC08020000 and No. XDC08020400) and the National Natural Science Foundation of China (No.61472393).}
\thanks{Jiwei Chen is with the Institute of Intelligent Machines, Hefei Institutes of Physical Science, Chinese Academy of Sciences, Hefei 230031, China, and also with University of Science and Technology of China, Hefei 230026, China, and Hefei University of Technology, Hefei 230009, China.}
\thanks{Kewei Wang is with the School of Computer Science, Faculty of Engineering, University of Sydney.}
\thanks{Wen Su is with the virtual reality laboratory in Zhejiang Sci-Tech University.}
\thanks{Zengfu Wang  is with the Institute of Intelligent Machines, Hefei Institutes of Physical Science, Chinese Academy of Sciences, Hefei 230031, China, and also with University of Science and Technology of China, Hefei 230026, China. (e-mail: zfwang@ustc.edu.cn).}}

\maketitle

\begin{abstract}
Crowd counting based on density maps is generally regarded as a regression task.Deep learning is used to learn the mapping between image content and crowd density distribution. Although great success has been achieved, some pedestrians far away from the camera are difficult to be detected. And the number of hard examples is often larger. Existing methods with simple Euclidean distance algorithm indiscriminately optimize the hard and easy examples so that the densities of hard examples are usually incorrectly predicted to be lower or even zero, which results in large counting errors. To address this problem, we are the first to propose the \textbf{H}ard \textbf{E}xample \textbf{F}ocusing (\textbf{HEF}) algorithm for the regression task of crowd counting. The HEF algorithm makes our model rapidly focus on hard examples by attenuating the contribution of easy examples. Then higher importance will be given to the hard examples with wrong estimations. Moreover, the scale variations in crowd scenes are large, and the scale annotations are labor-intensive and expensive. By proposing a multi-\textbf{S}cale \textbf{S}emantic \textbf{R}efining (\textbf{SSR}) strategy, lower layers of our model can break through the limitation of deep learning to capture semantic features of different scales to sufficiently deal with the scale variation. We perform extensive experiments on six benchmark datasets to verify the proposed method. Results indicate the superiority of our proposed method over the state-of-the-art methods. Moreover, our designed model is smaller and faster. 
\end{abstract}

\begin{IEEEkeywords}
Crowd counting, density map, multi-scale semantic refining strategy, hard example focusing.
\end{IEEEkeywords}

\section{Introduction}

Crowd counting based on computer vision can estimate the number of people in video frames or still images. It has attracted remarkable attention since its wide practical applications including public safety, urban planning, and traffic management \cite{zhang2015cross,idrees2013multi,babu2018divide}. To be specific, crowd density estimation can be utilized to monitor crowds in real time to prevent injuries caused by overcrowding. It can also provide advice on public resource allocation and real-time traffic dispatch. Moreover, the methods of crowd counting can be also extended in the cell counting, vehicle counting, and so on. However, due to the scale variation, high density, and non-uniform distribution in crowd scenes, this task is still challenging.

Great efforts have been devoted and various algorithms have been proposed in crowd counting. The algorithms are generally divided into detection and regression based methods. In detection based methods \cite{li2008estimating,lin2010shape,zeng2010robust}, pedestrians are detected and the number of positive detections denotes the crowd count. However, some pedestrians are too small to be detected in the dense crowd scene. In regression based methods \cite{chan2008privacy,chan2009bayesian,idrees2013multi}, the extracted features are mapped to the crowd counting result. However, the location of each human is omitted. In recent years, the convolutional neural network (CNN) based methods occupy an undisputed dominance. In CNN based methods \cite{zhang2016single,babu2018divide,CHEN2019}, the proposed models can automatically grasp the visual features and further encode them to a density map that records the quantity and location of pedestrians in images. They acquire the crowd counting result by summing up all values in the density map.


\begin{figure}[t]
\setlength{\abovecaptionskip}{0.5cm}
\setlength{\belowcaptionskip}{-0.9cm}
\begin{center}
   \includegraphics[width=1.0\linewidth]{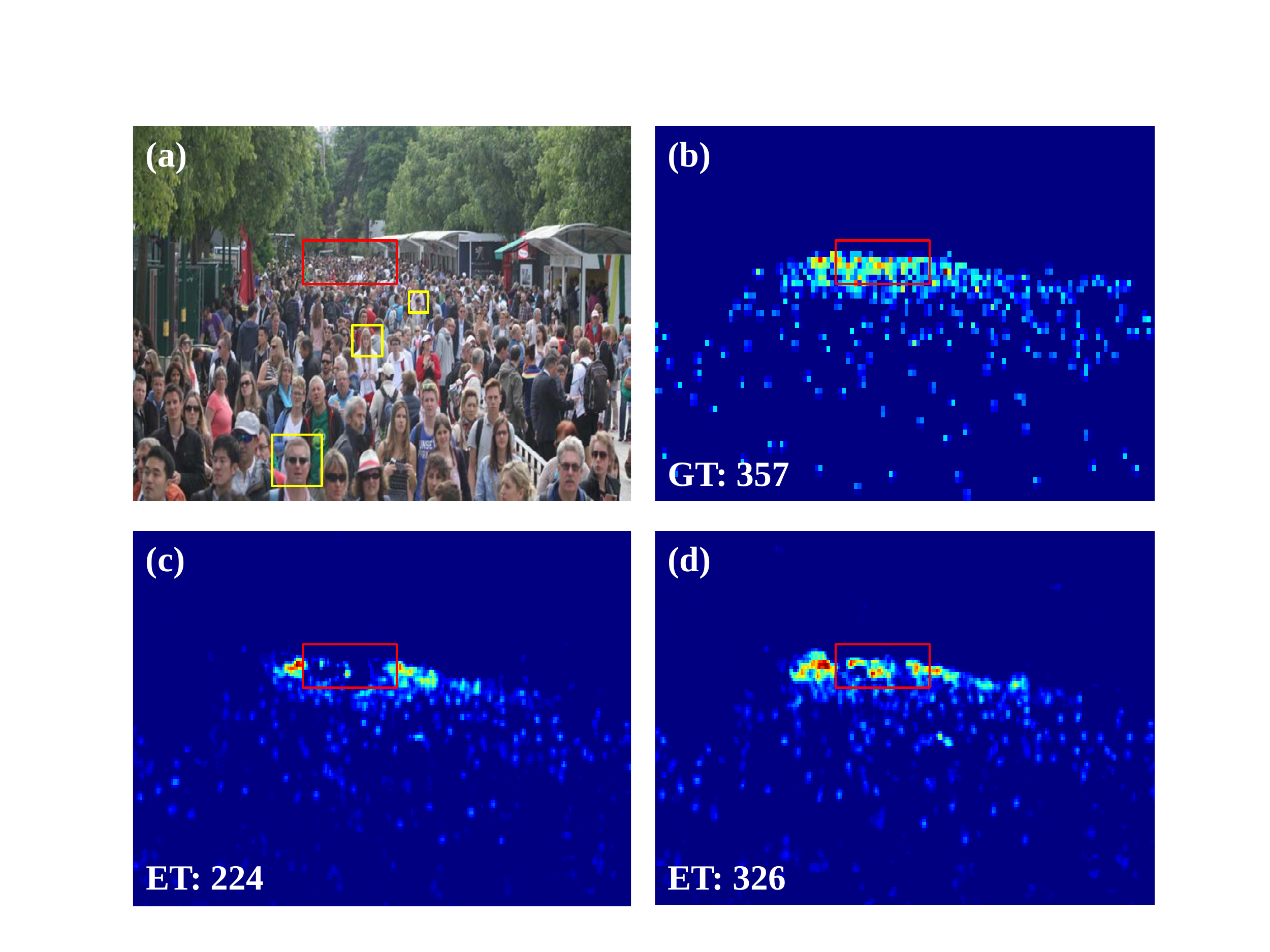}
   \caption{Examples of two issues in the crowd. 1) Hard example: pedestrians in red rectangles are difficult to be detected. 2) Scale variation: pedestrian head size in yellow rectangles of (a) varies with the distance from camera. (a) represents the input image. (b) represents the ground-truth density map. (c) represents the density map estimated by MCNN. (d) represents the density map estimated by MCNN with our HEF algorithm. GT represents the ground-truth count. ET represents the estimated count.}
\label{FIG:1}
\end{center}
\end{figure}

\begin{figure*}[t]
\begin{center} 
   \includegraphics[width=1.0\linewidth]{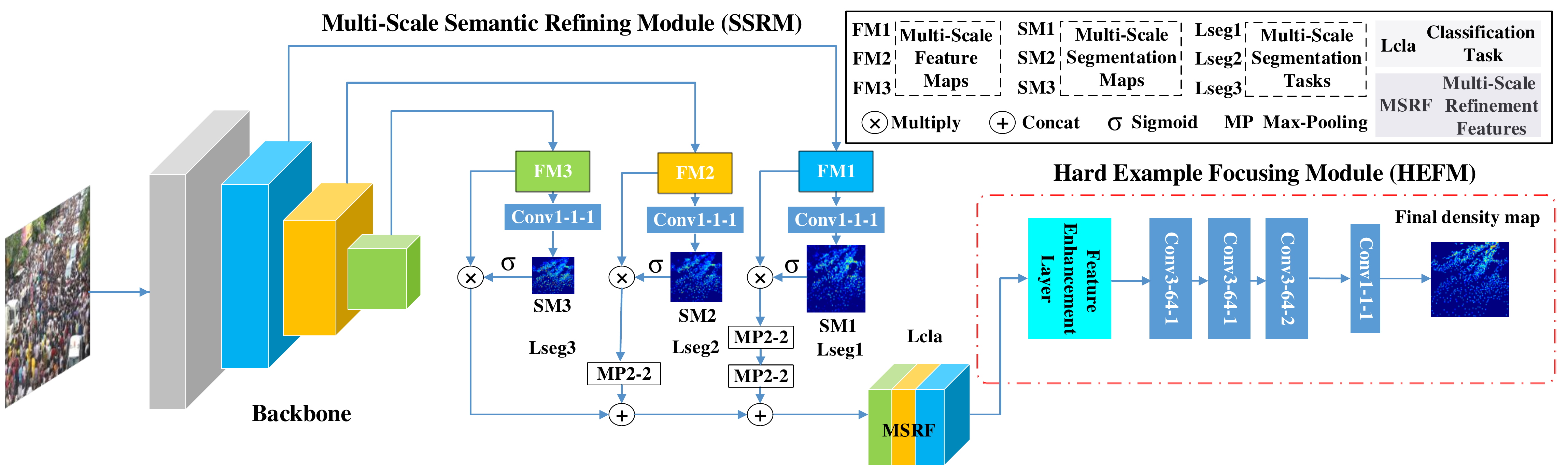}

   \caption{The architecture of the proposed SSR-HEF. The segmentation map of each scale is generated and mapped back to network to adaptively refine and extract multi-scale features in the SSRM. The hard example focusing task is completed in the HEFM. Its parameters of the convolutional layers are represented as “Con(kernel size)-(number of filters)-(dilation rate)”. Its parameters of the Max-Pooling layers are represented as “MP(kernel size)-(strides).}
\label{FIG:2}
\end{center}
\end{figure*}

In recent years, tremendous CNN models have been proposed to regress the crowd density distribution. Then the counting result can be obtained by integrating the final density map. Although they have achieved significant improvement in crowd counting, we find that the hard sample problem in crowd scenes has not been solved. As shown in Fig. \ref{FIG:1} (a), due to long distances from the camera, some pedestrians in red rectangles are difficult to be detected. And their densities are imprecisely predicted by MCNN \cite{zhang2016single} to be lower or even zero in Fig. \ref{FIG:1} (c). We think that existing methods with simple L2 loss give the same weight coefficient to each example and indiscriminately optimize them so that the hard examples aren't effectively focused on and optimized, which results in large counting errors.

 Although the focal loss \cite{lin2017focal} used in the classification task is proposed to successfully deal with the class imbalance, this algorithm can not be utilized in crowd counting because it is a regression task. There are essential differences between classification and regression tasks. The former is qualitative and the latter is quantitative. To address this issue, we are the first to propose a novel hard example focusing algorithm for the regression task of crowd counting. It can be used to tackle the hard example problem in crowd scenes. By applying this algorithm, higher importance will be given to the hard examples with wrong estimations. In principle, easy samples will be severely penalized and their weights will be attenuated exponentially, which in turn increases the importance of hard examples. From Fig. \ref{FIG:1} (d), we can observe that the MCNN with our HEF algorithm can effectively address the problem of hard examples and improve the accuracy of crowd counting.







Additionally, as illustrated in Fig. \ref{FIG:1} (a), pedestrian size in yellow rectangles varies with the distance from camera. Many CNN based methods \cite{Sindagi_2019_ICCV8,Jiang_2019_CVPR,Xiong_2019_ICCV} concatenate features from different depth layers of deep network to cope with the scale variation. However, as reported in \cite{10.1007/978-3-319-10590-1_53}, higher layers of deep CNN encode the semantic concepts, whereas lower layers capture primitive features. The semantic concepts can not be encoded in the lower layers of aforementioned methods. So negative concatenation of different depth features is not the best way to solve the problem of large scale variations. 

To sufficiently overcome the scale variation problem, we propose a multi-scale semantic refining strategy (as seen in Fig. \ref{FIG:2}). With the supervision of ground-truth segmentation of each scale, the multi-scale segmentation maps, where the highlighted places represent the crowd area, are generated and mapped back to network to adaptively refine the multi-scale features. Then the features with semantic concepts of different scales are further encoded to automatically contribute to generating the final robust density map. Unfortunately, the present crowd counting datasets do not provide the ground-truth segmentation of each scale. To obtain them without the substantial amount of extra manual labeling cost, we propose an automatic labeling method where the ground-truth segmentation is encoded by pasting the matrix of ones on the position of pedestrian head on a binary map, then the labels of multi-scale segmentation maps are obtained by a simple downsampling transformation.

Our contributions are summarized as follows:

  
  1.  We find the hard sample problem in the regression task of crowd counting. To address this problem, the weighted hard example focusing algorithm is proposed to make model focus on the hard samples with wrong estimations in the quantitative regression task.
  

 2. To sufficiently overcome the scale variation problem, we first propose a novel multi-scale semantic refining strategy. Multi-scale segmentation maps with semantic prior are proposed and mapped backed to network to adaptively refine and extract multi-scale features. This strategy also breaks through the limitation of deep learning to make lower layers of deep CNN capture semantic concepts. 

3. Extensive experiments are conducted on six benchmark datasets i.e. Shanghaitech Part\_A, Shanghaitech Part\_B, UCF\_CC\_50, UCF-QNRF, WorldExpo'10, and JHU-CROWD. Results demonstrate the superiority of our lightweight SSR-HEF over many state-of-the-art methods. Moreover, our designed model is smaller and faster.

In the remainder of this paper, we will discuss the related work in Section II. The detail of proposed method will be given in Section III. Experiment results with analysis will be shown in Section IV, followed by the conclusion in Section V.

\section{Related Work}

Recently, various algorithms have been proposed in crowd counting, especially for deep learning. We will introduce some of them.

\subsection{Detection based methods}
In preliminary researches, the detection framework is popular for crowd counting. Li et al. \cite{li2008estimating} detected the head of humans by Haar wavelets. Lin et al. \cite{lin2010shape} leveraged the part-template tree to detect and segment the crowd. Zeng et al. \cite{zeng2010robust} utilized the HOG and LBP to detect key parts of human body. They get the crowd counting result by summing the number of positive detections. However, they may not generalize to dense crowds because of some too small humans.

\subsection{Regression based methods}
The regression based methods are more robust than the detection based methods in dense crowd scenes. They extract some pivotal features, such as edge, texture, and segmentation features, from the input image and further map them to the crowd counting result without detection. In \cite{chan2008privacy}, the crowd count was regressed by the Gaussian model. In \cite{chan2009bayesian}, the Poisson model was reported to map extracted features to the crowd counting result. In \cite{idrees2013multi}, the features of human heads were employed to count the crowd. However, they lost the location of the crowd.


\subsection{CNN based methods}

Deep learning has obtained commendable performance in many fields. And it improves the accuracy of crowd counting dramatically. The CNN based methods \cite{zhang2016single,chen2021crowd,abousamra2021localization} regress to the density map where the location and quantity of crowd are recorded. Due to the serious scale variation in the crowd, many methods \cite{zhang2016single,babu2018divide,Jiang_2019_CVPR,Shi_2019_CVPR,Yang_2020_CVPR} are proposed to overcome it. In \cite{zhang2016single}, the MCNN with three-branch CNNs was designed. In \cite{babu2018divide}, the IG-CNN could adapt the scale variation by increasing its branches. Multi-scale feature maps of different depth layers were concatenated in \cite{Jiang_2019_CVPR}. PACNN \cite{Shi_2019_CVPR} introduced the perspective map into density regression. MBTTB-SCFB \cite{Sindagi_2019_ICCV8} fused the bottom-top and top-bottom features. Yang et al. \cite{Yang_2020_CVPR} proposed a reverse perspective network to evaluate and correct the distortions of input images; In \cite{Wang_2019_CVPR}, Wang et al. used game characters to synthesize plenty of data to train models to mitigate the overfitting caused by insufficient data. In \cite{Wan_2019_CVPR}, the semantic prior was exploited to contribute to crowd counting. In \cite{Xiong_2019_ICCV}, the S-DCNet trained on closed sets could generalize well to open sets. In \cite{2020Adaptive}, a local counting map was proposed to reduce the inconsistency between training targets and evaluation metrics. In \cite{abousamra2021localization}, a topological method was proposed to reduce the error of crowd localization.

\begin{figure}[t]
\setlength{\abovecaptionskip}{0.567cm}
\setlength{\belowcaptionskip}{-0.28cm}
\begin{center}
   \includegraphics[width=1.0\linewidth]{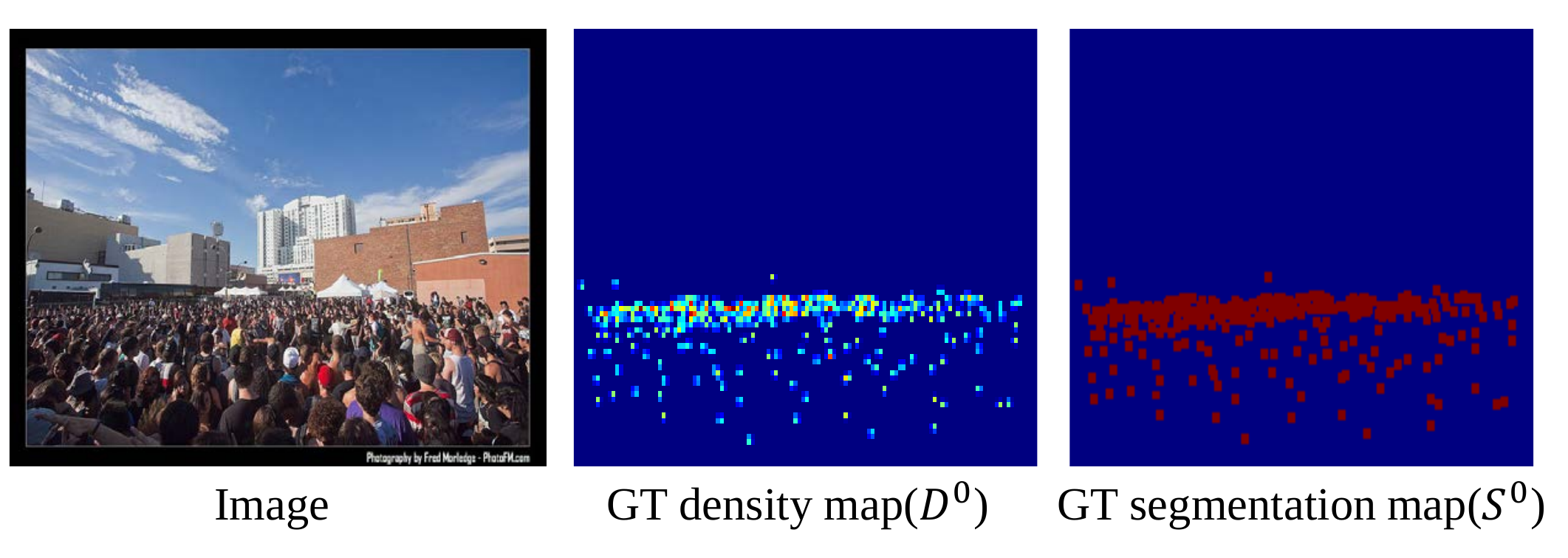}
   \caption{Some visualization results of ground truth.}
\label{FIG:4}
\end{center}
\end{figure}

In \cite{liu2021cross}, the optical and thermal information was mined to recognize pedestrians. In \cite{Zhang_2021_CVPR}, a cross-view cross-scene multi-view counting method was proposed to improve the performance and generalization of the model. In \cite{Wan_2021_CVPR}, a generalized loss function based on an unbalanced optimal transport was introduced for crowd counting and localization. In \cite{Song_2021_ICCV}, a purely point-based framework with a new metric was designed for joint crowd counting and individual localization. In \cite{Meng_2021_ICCV}, a spatial uncertainty-aware semi-supervised method with regularized surrogate task was proposed to reduce the annotation cost. In \cite{Chen_2021_ICCV}, the domain-specific knowledge propagating model was constructed to unbiasedly learn the knowledge from multi-domain data at the same time. In \cite{DBLPjournals/corr/abs-2012-00452}, the temporal consistency was explored to estimate crowd flows and infer crowd densities. In \cite{9346018}, a locality-aware data partition method was proposed to tackle the severe under-estimation and over-estimation problems. 

Although they have achieved great success in crowd counting, we notice that all of them do not find the problem of hard examples in crowd counting. To the best of our knowledge, we are the first to propose the novel hard example focusing model to address this problem. In addition, lower layers of all the aforementioned methods only capture primitive features and can not encode semantic concepts in deep CNN. We propose a multi-scale semantic refining strategy to break the limitation and sufficiently overcome the scale variation. The multi-layer convolution feature fusion \cite{2018Object} and feature pyramid fusion \cite{2020Mask} are effective in dealing with the large scale variation. Moreover, the features with high discrimination \cite{2017Dual} and low correlation \cite{2013PalmHash} can be selected and provided with high weights in fusion, which inspires us to further explore the multi-scale feature fusion with adaptive weighting to cope with the scale variation in future works.


\section{Proposed Method}
In this paper, we propose a multi-\textbf{S}cale \textbf{S}egmentation \textbf{R}efining and \textbf{H}ard \textbf{E}xample \textbf{F}ocusing model named SSR-HEF. It is a novel lightweight CNN model. An overview of our SSR-HEF is shown in Fig. \ref{FIG:2}. It consists of two modules: Multi-Scale Segmentation Refining Module (SSRM) and Hard Example Focusing Module (HEFM). In the SSRM, multi-scale segmentation maps are encoded and mapped backed to network to adaptively refine and extract multi-scale features. In the HEFM, the hard examples are focused on automatically and the final highly refined density map is generated. They are elaborated as follows.

\subsection{Ground Truth Generation}

The head position $p$ of each human is annotated in crowd counting dataset. Following the method reported in \cite{sindagi2017cnn}, we utilize the Gaussian kernel centered on each head position $p$ to encode the ground-truth density map. To ensure that the actual crowd counting result can be obtained by integrating each value over the entire ground-truth density map, the Gaussian kernel has been normalized to 1. The density at any pixel position $a$ in the ground-truth density map $D$ is formulated by Eq.(1):

\begin{equation}
 D(a)=\sum _{p\in P}N(a;\mu,\beta),
\end{equation}
where $P$ represents the complete set of annotated points in each image. $N(a;\mu,\beta)$ represents the Gaussian kernel with mean $\mu$ and variance $\beta$.

\begin{figure}[t]
\setlength{\abovecaptionskip}{0.567cm}
\setlength{\belowcaptionskip}{-0.28cm}
\begin{center}
   \includegraphics[width=1.0\linewidth]{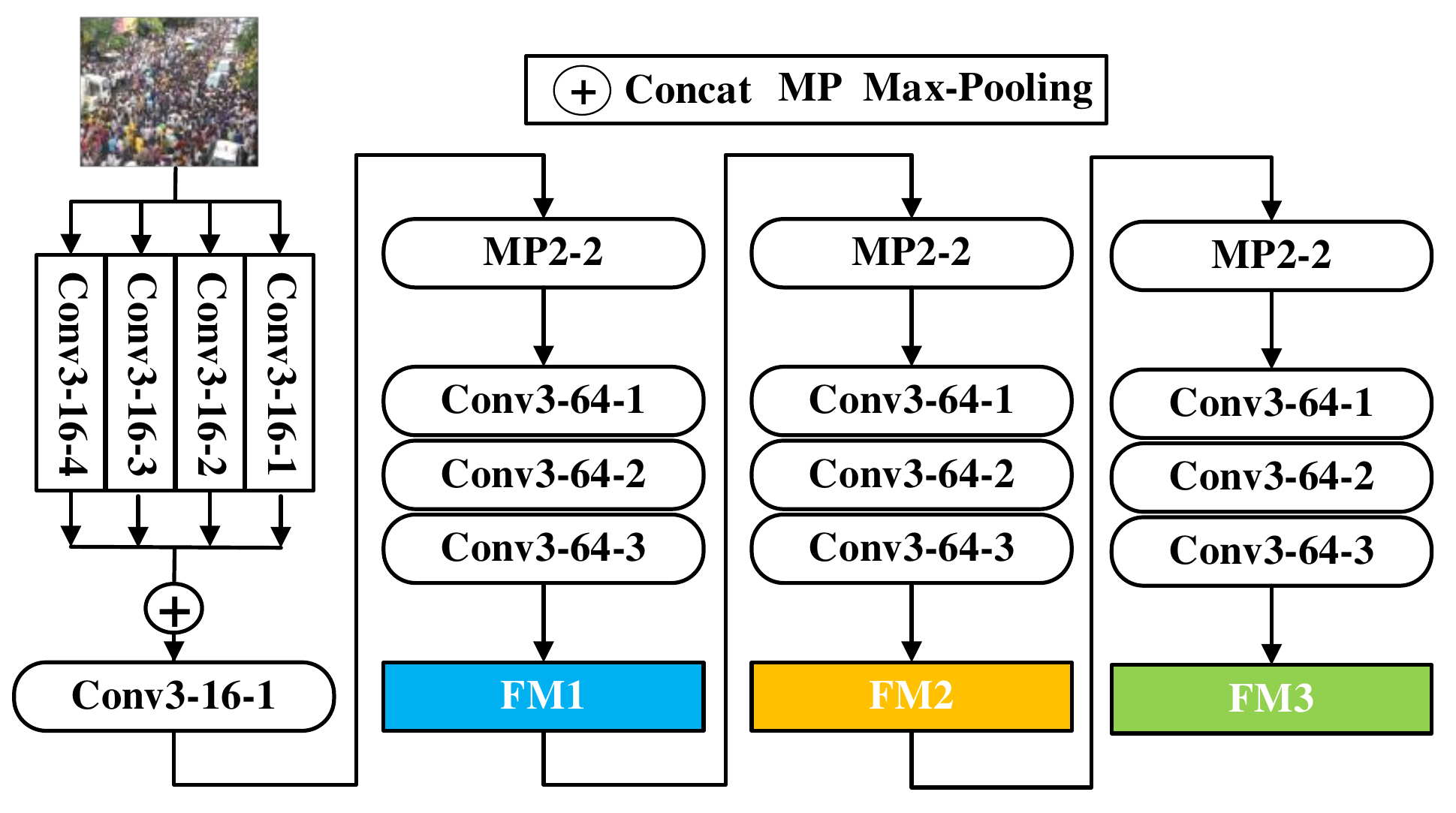}
   \caption{The configurations of designed backbone in the SSRM. Its parameters of convolutional layers are represented as “Con(kernel size)-(number of filters)-(dilation rate)”. Its parameters of the Max-Pooling layers are represented as “MP(kernel size)-(strides).}

\label{FIG:3}
\end{center}
\end{figure}

Due to the large and dense crowd in datasets, it is expensive to obtain multi-scale segmentation labels by manual labeling. For example, ShanghaiTeach Part\_A needs 241,677 annotations. Hence, an automatic labeling method is proposed to encode multi-scale segmentation labels. Specifically, as with the input to generate the ground-truth density map, the position $p$ of each head is annotated and given. The ground-truth segmentation map $S^{0}$ with the size of original image can be encoded by pasting the ones template on a binary map, and each ones template is centered on each head position $p$. The size of ones template is 15$\times$15 and values in the ones template are all 1. The output is the ground-truth segmentation map $S^{0}$ and it is a binary map. Then multi-scale segmentation labels $S^{1/2},S^{1/4},S^{1/8}$ can be obtained through the following two steps. Firstly, $S^{0}$ is the input and the multi-scale feature maps (MSFM) with the sizes of 1/2, 1/4, 1/8 of original image are resized from $S^{0}$. Secondly, these multi-scale segmentation labels $S^{1/2},S^{1/4},S^{1/8}$ are encoded from MSFM by the binarization where if the pixel value is greater than 0, it is set to 1, otherwise, it is set to 0. The outputs are multi-scale segmentation labels $S^{1/2},S^{1/4},S^{1/8}$ and they are also the binary map. Some visualization results are given in Fig. \ref{FIG:4}.

\subsection{Multi-Scale Semantic Refining Module}
As shown in Fig. \ref{FIG:2}, this module is first proposed to sufficiently cope with the scale variation problem by mapping predicted multi-scale segmentation maps back to network to adaptively refine and extract multi-scale features, which enhances the scale encoding ability of the network. The configurations of designed backbone in this module are shown in Fig. \ref{FIG:3}. To improve the robustness of our model to the scale variation, four different dilation convolution scales are utilized to encode the input image at the beginning of this backbone. Every convolution is followed by ReLU. The dilation convolutions are widely used to broaden receptive fields with fewer parameters, which can also avoid overfitting. Additionally, three max-pooling layers (size 2$\times$2, strides 2) are exploited to extract multi-scale features.


\begin{table}[]
\setlength{\abovecaptionskip}{-0.15cm}
\setlength{\belowcaptionskip}{-8cm}
\caption{The configurations of the classification task.}
\label{Table:1}
\begin{center}
\begin{tabular}{|c|c|c|c|}
\hline
\multicolumn{4}{|c|}{Input (MSRF)}                                                                                                  \\ \hline
\multicolumn{4}{|c|}{Spatial Pyramid Pooling (16$\times$16)}                                                                                          \\ \hline
\multicolumn{4}{|c|}{FC-64}                                                                                             \\ \hline
\multicolumn{4}{|c|}{FC-15}                                                                                            \\ \hline


\end{tabular}

\end{center}
\end{table}

The detailed structure of multi-scale semantic refinement is illustrated in Fig. \ref{FIG:2}. With the supervision of multi-scale segmentation labels, the multi-scale segmentation maps where the highlighted places represent the crowd area are generated and lower layers of the network can capture semantic concepts. To adaptively refine multi-scale features, after the sigmoid layer $\sigma$, the segmentation maps with semantic prior are mapped back to the output features (denoted by $F_{m}^{h\times w\times c}$) of different depth layers by performing an element-wise multiplication i.e., $\hat{F}_{m}^{h\times w\times c}=F_{m}^{h\times w\times c}\bigodot\sigma(\hat{S}_{m})$. Then they are further fused to obtain multi-scale semantic refining features (MSRF). The Dice loss \cite{milletari2016v} is utilized to optimize these segmentation tasks. Mathematically,

\begin{center}
\begin{equation}
L_{segs} = \sum_{m=1}^{M}[1-\frac{2\hat{S}_{m}S_{m}}{\hat{S}_m^{2}+S_m^{2} }],
\end{equation}

\end{center}where $M$ represents the number of scales and it is set to 3 in this paper. $S_{m}$ represents the ground-truth segmentation map of each scale which respectively corresponds to $S^{1/2}, S^{1/4}, S^{1/8}$ encoded in the Ground Truth Generation. $\hat{S}_{m}$ represents the predicted segmentation map of each scale. Although \cite{2019Leveraging,2019PCC} also use segmentation tasks, there are essential differences. \cite{2019Leveraging} treats it as an auxiliary multi-task learning, and the segmentation task is used to distinguish the foreground and background in \cite{2019PCC}. While we propose the multi-scale semantic refining strategy to explicitly solve multi-scale problems and also break through the limitation of deep learning to make lower layers of deep CNN capture some semantic concepts.


To incorporate the global-level cue into the MSRF to contribute in encoding the final highly refined density map, the auxiliary classification task \cite{sindagi2017cnn} divides these crowd counts into different classes to give an overall rough estimation of the total population in the image. In contrast to \cite{sindagi2017cnn} where the classification task with an additional CNN branch adds a lot of redundant parameters, our classification task directly shares the backbone with other tasks, which reduces the redundancy of our model and makes our model easier to train. Additionally, our SSR-HEF solves the scale variation and hard example problem for crowd counting. The configurations of the classification task are given in Table \ref{Table:1}. The cross-entropy loss is utilized to optimize this task. 


\begin{equation}
L_{cla} = -\frac{1}{N}\sum ^{N}_{n=1}\sum ^{K}_{k=1}\left [ \left ( y^{k}_{n} \right )log (\hat{y}_{n}) \right ],
\end{equation}where $N$ represents the number of test samples. $\hat{y}_{n}$ represents the output of classification network. $y^{k}_{n}$ represents the actual class of crowd counts. $K$ represents the classes of crowd counts and it is set to 15 in this paper. Since the populations in different images of different datasets vary greatly, we propose the adaptive classification criteria to encode the labels of classification tasks for different datasets. Specifically, we obtain the maximum count of people $C$ in the dataset, and we use $Thr= C / K$ as the adaptive threshold to divide these crowd counts of the dataset into different classes. A simple example is given to illustrate the encoding process of the actual class of crowd counts. For instance, if the number of people in a dataset ranges from 0 to 30, and we quantify these crowd counts of this dataset into three classes. $Thr= 10$. And the images of populations between 0 and 10 fall into the first class. The images of populations between 10 and 20 fall into the second class. And the images of populations between 20 and 30 fall into the third class. $Thrs$ of different datasets are given in Table \ref{Table:2}.

\begin{figure}[t]
\setlength{\abovecaptionskip}{-0.067cm}
\setlength{\belowcaptionskip}{-0.6cm}
\begin{center}
   \includegraphics[width=1.0\linewidth]{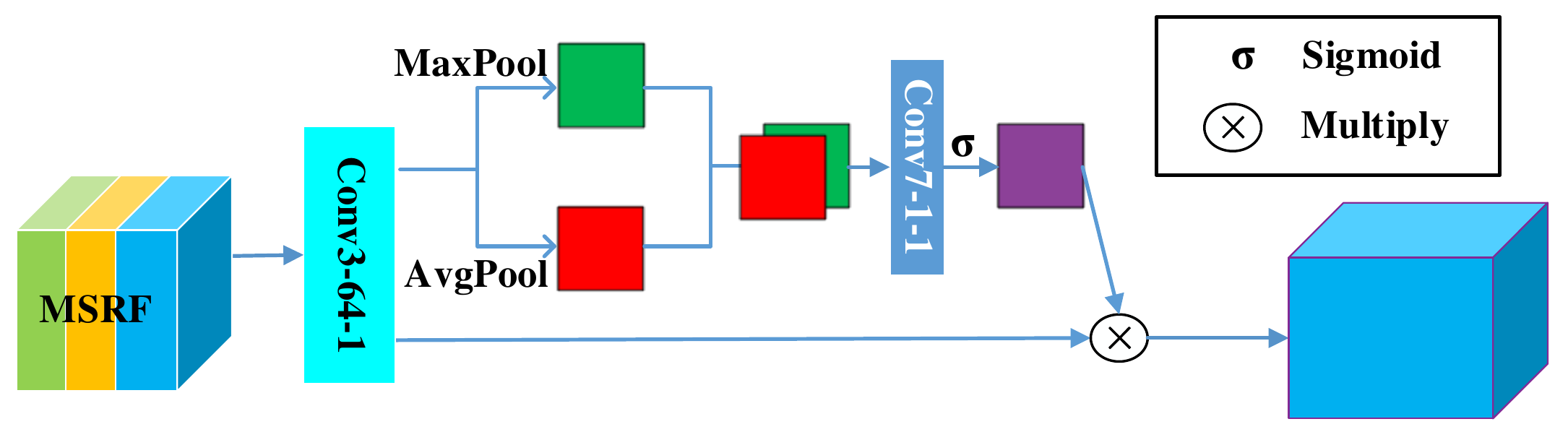}

   \caption{The architecture of the feature enhancement layer. Its parameters of convolutional layers are represented as “Con(kernel size)-(number of filters)- (dilation rate)”.}
\label{FIG:5}

\end{center}

\end{figure}

\begin{table*}
\setlength{\abovecaptionskip}{0.067cm}
\setlength{\belowcaptionskip}{-0.58cm}
\begin{center}
\caption{Statistics of these datasets: Ave Resolution denotes the average resolution of images. Color denotes the color of images. Num Images denote the number of images in the dataset; Max Count denotes the maximum crowd counts in the dataset; Classification Thr denotes the threshold for classifying crowd counts of the dataset; Ave count denotes the average crowd counts in the dataset; Annotations denote the total number of labeled people in the dataset.}
\label{Table:2}
\begin{tabular}{|c|c|c|c|c|c|c|c|c|}\hline
Dataset                                &Ave Resolution & Color & Num Images & Max Count &Classification Thr & Ave Count & Annotations \\ \hline 

ShanghaiTech Part\_A \cite{zhang2016single}          & 589 $\times$ 868   & RGB,Grey  & 482  & 3139 & 209.3 & 501.4   & 241,677 \\  \hline 
ShanghaiTech Part\_B \cite{zhang2016single}          & 768 $\times$ 1024   & RGB & 716  & 578 &38.5  & 123.6   & 88,488 \\  \hline 
UCF\_CC\_50 \cite{idrees2013multi}               & 2101 $\times$ 2888   & Grey  & 50   & 4543 & 302.9  & 1279.5   & 63,974 \\ \hline 
UCF-QNRF \cite{Idrees_2018_ECCV}     & 2013 $\times$ 2902  & RGB,Grey  & 1535  & 12865 & 857.7  & 815.4   & 1,251,642 \\  \hline 
WorldExpo’10 \cite{zhang2015cross}          & 576 $\times$ 720   & RGB  & 3980  & 253 & 16.9  & 50.2   & 199,923 \\ \hline
JHU-CROWD \cite{Sindagi_2019_ICCV}          & 1450 $\times$ 900    & RGB,Grey  & 4250  & 7286 & 485.7  & 262.3   & 1,114,785 \\ \hline  
\end{tabular}
\end{center}
\end{table*}


\subsection{Hard Example Focusing Module}

Strong crowd characteristics are crucial for counting. The Feature Enhancement Layer \cite{woo2018cbam} is used to strengthen the crowd feature representations for the Hard Example Focusing task where the counting result is obtained.

{\bfseries Feature Enhancement Layer:}
This layer is included in the HEFM. Details of this layer are illustrated in Fig. \ref{FIG:5}. The outputs of SSRM are fed into this layer. The convolution layer followed by ReLU is employed to encode them. Then the max pool and average pool along the channel axis are utilized to respectively extract the highlighted representations from the feature maps. After concatenation, these highlighted representations followed by convolution and sigmoid layer are further encoded and mapped back to network to enhance the feature responses of the crowd.


\begin{figure}[t]
\setlength{\abovecaptionskip}{0.067cm}
\setlength{\belowcaptionskip}{-0.18cm}
\begin{center}
   \includegraphics[width=1.0\linewidth]{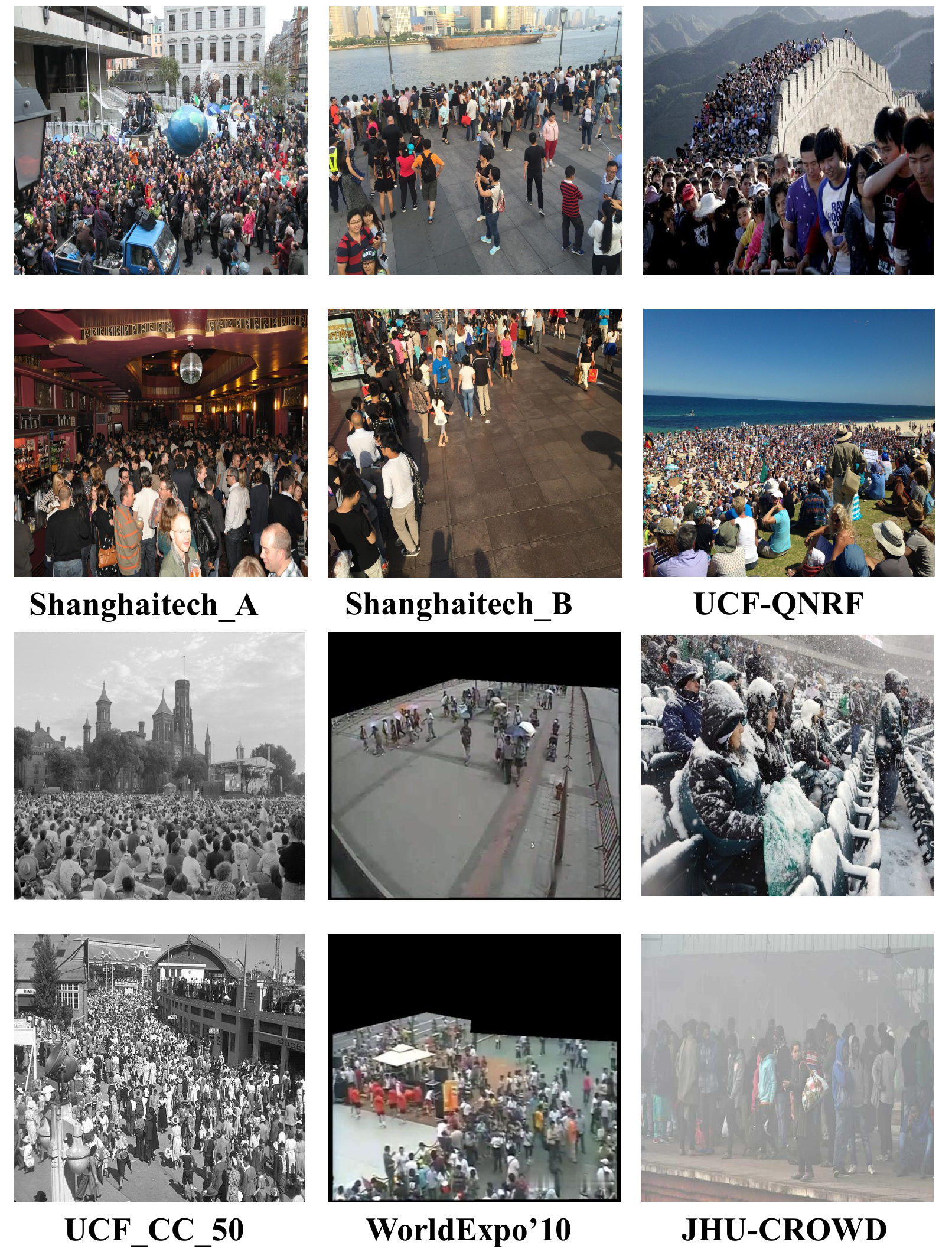}
   \caption{Some examples of these datasets.}

\label{FIG:6}
\end{center}
\end{figure}

{\bfseries Hard Example Focusing:}
We first find the hard example problem in the regression task of crowd counting. Existing methods with L2 loss give the same weight coefficient to each example and indiscriminately optimize them. The hard examples have not received effective focus and optimization. So the density of them is usually predicted incorrectly, which often shows that the estimated density is lower than the density of ground truth. Inspired by the focal loss \cite{lin2017focal} used in the classification task, we propose a novel HEF algorithm which can be directly used in the regression task of crowd counting to address this problem. By using this HEF algorithm, the hard examples with wrong estimations can be rapidly focused on. In principle, the weight modulating factor reduces the contribution of L2 loss of easy examples, which in turn increases the importance of hard examples. More formally, the proposed algorithm with Euclidean distance is utilized as a loss function to optimize the final density map. 


\begin{multline}
\begin{aligned}
L_{hef} =\frac{1}{W*H}\sum ^{W}_{w=1}\sum ^{H}_{h=1} \quad \quad \quad \quad \quad \quad \\
(1-\sigma (D\left ( X_{(w,h)}; \theta \right )))^{\gamma} \left \| D\left ( X_{(w,h)}; \theta \right ) - D_{(w,h)} \right \|^{2}_{2},
\end{aligned}
\end{multline}
where $W$ and $H$ represent the width and height of estimated density map, respectively. $\sigma$ represents the sigmoid function. $D\left ( X_{(w,h)}; \theta \right )$ represents the density of position $(w,h)$ in the final estimated density map. $\theta$ represents the parameters of network. And all values in the final estimated density map are summed to acquire the crowd counting result. $D_{(w,h)}$ represents the density in the ground-truth density map. As recommended by \cite{lin2017focal}, exponent $\gamma$ is set to 2. The weight modulating factor $(1-\sigma (D\left ( X_{(w,h)}; \theta \right )))^{\gamma}$ is introduced to the Euclidean distance to attenuate the contribution from easy examples. For instance, when the hard sample isn’t detected correctly and its density was incorrectly predicted to be lower or even 0, the weight modulating factor tends to 1, and the loss is almost unaffected. For the easy sample, its density is easy to estimate, and the weight modulating factor tends to 0, then its loss is down-weighted. In this way, its weights will be greatly reduced during backpropagation, which in turn increases the importance of hard samples.

 In \cite{Sindagi_2019_ICCV}, the L2-loss reweighted with uncertainty is proposed for crowd counting. This work is different from ours. The L2-loss reweighted with uncertainty is designed to increase the effectiveness of the residual learning, where only highly confident residuals are allowed to get propagated to the output. Our HEF algorithm with the focal factor is proposed to focus on hard examples of the crowd, where hard examples can be focused on by attenuating the contribution of easy examples. In \cite{2017Simple}, the cross-modal learning to rank with the self-paced learning theory is proposed to learn the samples from simple to complex, which points us to a new direction for hard examples mining in crowd counting.



\section{Experiments}

 With the development of crowd counting, more than 20 datasets have been proposed. And extensive experiments are separately conducted on six offline benchmark datasets including Shanghaitech Part\_A \cite{zhang2016single}, Shanghaitech Part\_B \cite{zhang2016single}, UCF\_CC\_50 \cite{idrees2013multi}, UCF-QNRF \cite{Idrees_2018_ECCV}, WorldExpo'10 \cite{zhang2015cross}, and JHU-CROWD \cite{Sindagi_2019_ICCV}.  Statistics of them are elaborated in Table \ref{Table:2}. Some representative examples of these datasets are shown in Fig. \ref{FIG:6}.  We also conducted experiments on the online dataset (NWPU-Crowd \cite{gao2020nwpu}). Real-time results can be observed at \href{https://www.crowdbenchmark.com/nwpucrowd.html}{https://www.crowdbenchmark.com/nwpucrowd.html}. To alleviate overfitting caused by limited training data, patches with the sizes of 1/16, 1/4, 1 of original image are randomly cropped from every image. The numbers of patches in each size are set to 9, 4, and 1, respectively. Meanwhile, random noise and flipping are also employed on them with a probability of 0.5 during the training process. The learning rate is set as 3e-5. And the Adam optimization with a batch size of 1 is used to optimize the overall loss function in an end-to-end manner. The overall loss is given as follows:
\begin{center}
\begin{equation}
 L_{overall} =L_{hef} +  \lambda_{1}L_{segs} + \lambda_{2}L_{cla},
\end{equation}
\end{center} during the tuning process, we find that $L_{segs}$, $ L_{cla} $ are large and overwhelm the losses $ L_{hef} $ in the hard example focusing task. To balance the magnitude of different tasks, $\lambda_{1}$ is set to 1e-2 and $\lambda_{2}$ is set to 1e-3. 


\subsection{Evaluation Metric}

 Following existing standard metrics in \cite{zhang2016single,Wang_2019_CVPR,Yang_2020_CVPR}, the Mean Absolute Error (MAE) and Mean Squared Error (MSE) are leveraged to evaluate the performance of different methods, which are defined as follows: 
\begin{center}
\begin{equation}
MAE = \frac{1}{E}\sum _{e=1}^{E}\left | z_{e}-\widetilde{z_{e}} \right |, MSE = \sqrt{\frac{1}{E}\sum ^{E}_{e=1}\left ( z_{e} - \widetilde{z_{e}} \right )^{2}}.
\end{equation}
\end{center}



Where $E$ represents the total number of test images, and $z_{e}$ represents the actual crowd count by summing each value in the ground-truth density map. $\widetilde{z_{e}}$ represents the estimated crowd count by summing each value in the final density map. MAE represents the accuracy of crowd counting and MSE represents the robustness of proposed methods.

\begin{table*}[]
\caption{Estimation errors on the WorldExpo'10 dataset.}
\label{Table:3}
\begin{center}
\begin{tabular}{|p{1cm}|c|c|c|c|c|c|c|}\hline
\multicolumn{1}{|c|}{Method}                                & Scene 1 & Scene 2 & Scene 3 & Scene 4 & Scene 5 & \textcolor{green}{Average} \\ \hline
\hline
\multicolumn{1}{|c|}{LBP+RR}                & 13.6   & 58.9  &37.1   & 21.8  & 23.4   & \textcolor{green}{31.9} \\ \hline
\multicolumn{1}{|c|}{Wang et al. \cite{Wang_2019_CVPR}}         & 4.3   & 59.1  & 43.7  & 17.0  & 7.6   & \textcolor{green}{26.3} \\ \hline

\multicolumn{1}{|c|}{IG-CNN \cite{babu2018divide}}  & 2.6   & 16.1  & 10.2  & 20.2  &7.6  &\textcolor{green}{11.3} \\ \hline  



\multicolumn{1}{|c|}{Wan et al. \cite{Wan_2019_CVPR}}                 & 2.9   & 15.0  & \textcolor{blue}{\textbf{7.2}}   & 14.7  & 2.6   & \textcolor{green}{8.5}  \\ \hline


\multicolumn{1}{|c|}{Yang et al. \cite{Yang_2020_CVPR}}                & 2.4   & 10.2  & 9.7   & 11.5  & 3.8   & \textcolor{green}{8.2}  \\ \hline
\multicolumn{1}{|c|}{TEDnet \cite{Jiang_2019_CVPR}}                 & 2.3   & 10.1  & 11.3   & 13.8  & 2.6   & \textcolor{green}{8.0}  \\ \hline


\multicolumn{1}{|c|}{PACNN \cite{Shi_2019_CVPR}}               & 2.3   & 12.5  & 9.1   &  11.2  & 3.8   & \textcolor{green}{7.8}  \\ \hline






\multicolumn{1}{|c|}{SSR-HEF(OURS)}                                 & \textcolor{blue}{\textbf{1.9}}   & \textcolor{blue}{\textbf{9.8}}  & 8.4   & \textcolor{blue}{\textbf{10.0}}  & \textcolor{blue}{\textbf{2.4}}   &\textcolor{blue}{\textbf{6.5}}  \\ \hline
\end{tabular}
\end{center}
\end{table*}

\begin{table*}
\begin{center}
\centering
\caption{Estimation errors on the JHU-CROWD dataset.}

\label{Table:61}
\begin{tabular}{|c|c|c|c|c|c|c|c|c|c|c|c|c|}
\hline

\multicolumn{1}{|c|}{\multirow{1}{*}{Category}}  & \multicolumn{2}{c|}{Distractors}   & \multicolumn{2}{c|}{Low}  & \multicolumn{2}{c|}{Medium}  & \multicolumn{2}{c|}{High} & \multicolumn{2}{c|}{Weather} & \multicolumn{2}{c|}{Overall} \\ \hline  
\multicolumn{1}{|c|}{\multirow{1}{*}{Methods}}                & MAE           & MSE              & MAE           & MSE       & MAE           & MSE    & MAE         & MSE           & MAE           & MSE        & MAE           & MSE        \\  \hline
MCNN \cite{zhang2016single}            & 103.8                  & 238.5                   & 37.7                   & 92.5        & 84.1             & 185.2        & 499.6              &795.5           & 128.2       & 288.3   & 109.3    & 291.0    \\  \hline

CMTL \cite{sindagi2017cnn}        & 138.5                   & 263.8            & 47.0                       & 106.0          & 82.4                  & 198.3           & 407.8             & 660.2           & 117.8      & 260.1  & 102.5    & 262.6      \\  \hline

Switching CNN \cite{sam2017switching}      & 100.5             & 235.5      & 32.1            & 80.5            & 76.1                 & 173.1           & 395.1             & 640.1           & 105.1         & 245.2   & 99.1        & 255.1      \\  \hline

SA-Net \cite{cao2018scale}       & 71.9             & 167.7                       & 30.0                & 76.6            & 65.4                 & 121.5           & 516.3             & 762.7          & 99.4            & 234.9   & 98.0        & 260.3      \\  \hline

CSR-Net \cite{li2018csrnet}      & 44.3             & 102.4                       & 15.8                & 39.9            & 48.4                 & 77.7           & 463.5             & 746.1           & 96.5          & 284.6   & 78.4      & 242.7      \\  \hline

CG-DRCN \cite{Sindagi_2019_ICCV}      & 43.4             & 97.8                       & 15.7                & 38.9            & 44.0                & 73.2           & 346.2             & 569.5           & 80.9          & 227.3   & 66.1      & 195.5      \\  \hline

SSR-HEF & \textcolor{blue}{\textbf{32.2}}    &\textcolor{blue}{\textbf{71.8}}  & \textcolor{blue}{\textbf{11.6}}    & \textcolor{blue}{\textbf{20.7}}  & \textcolor{blue}{\textbf{34.8}}  & \textcolor{blue}{\textbf{51.9}} &\textcolor{blue}{\textbf{268.3}}  &\textcolor{blue}{\textbf{378.6}}  &\textcolor{blue}{\textbf{69.5}}    &\textcolor{blue}{\textbf{120.1}}  &\textcolor{blue}{\textbf{51.3}}    &\textcolor{blue}{\textbf{101.6}}\\ \hline

\end{tabular}
\end{center}

\end{table*}

\begin{table}[]
\begin{center}
\centering
\caption{Estimation errors on the ShanghaiTeach.}
\label{Table:4} 
\begin{tabular}{|c|c|c|c|c|}
\hline
\multicolumn{1}{|c|}{\multirow{2}{*}{Methods}}                                                & \multicolumn{2}{c|}{Part\_A}   & \multicolumn{2}{c|}{Part\_B}  \\ \cline{2-5}
                               & \multicolumn{1}{c|}{MAE}            & \multicolumn{1}{c|}{MSE}   & \multicolumn{1}{c|}{MAE}            & \multicolumn{1}{c|}{MSE}            \\ \hline
\hline
\multicolumn{1}{|c|}{LBP + RR }                & \multicolumn{1}{c|}{303.2}          & \multicolumn{1}{c|}{371.0}  & \multicolumn{1}{c|}{59.1}          & \multicolumn{1}{c|}{81.7}     \\ \hline
\multicolumn{1}{|c|}{Wang et al. \cite{Wang_2019_CVPR}}                & \multicolumn{1}{c|}{123.4}          & \multicolumn{1}{c|}{193.4}  & \multicolumn{1}{c|}{19.9}          & \multicolumn{1}{c|}{28.3}       \\ \hline

\multicolumn{1}{|c|}{IG-CNN \cite{babu2018divide}}        & \multicolumn{1}{c|}{72.5}          & \multicolumn{1}{c|}{118.2} & \multicolumn{1}{c|}{13.6}          & \multicolumn{1}{c|}{21.1}         \\ \hline           




\multicolumn{1}{|c|}{TEDnet \cite{Jiang_2019_CVPR}}        & \multicolumn{1}{c|}{64.2}          & \multicolumn{1}{c|}{109.1}   & \multicolumn{1}{c|}{8.2}          & \multicolumn{1}{c|}{12.8}          \\ \hline





\multicolumn{1}{|c|}{PACNN \cite{Shi_2019_CVPR}}        & \multicolumn{1}{c|}{62.4}          & \multicolumn{1}{c|}{102.0}  & \multicolumn{1}{c|}{8.9}        & \multicolumn{1}{c|}{13.5}          \\ \hline

\multicolumn{1}{|c|}{AMRNet \cite{2020Adaptive}}        & \multicolumn{1}{c|}{61.6}          & \multicolumn{1}{c|}{98.4}   &\multicolumn{1}{c|}{7.0}          & \multicolumn{1}{c|}{11.0}           \\ \hline
\multicolumn{1}{|c|}{Wan et al. \cite{Wan_2019_CVPR}}        & \multicolumn{1}{c|}{61.3}          & \multicolumn{1}{c|}{101.4}  & \multicolumn{1}{c|}{8.7}          & \multicolumn{1}{c|}{13.6}          \\ \hline
\multicolumn{1}{|c|}{Yang et al. \cite{Yang_2020_CVPR}}        & \multicolumn{1}{c|}{61.2}          & \multicolumn{1}{c|}{96.9}  & \multicolumn{1}{c|}{8.1}          & \multicolumn{1}{c|}{11.6}          \\ \hline

\multicolumn{1}{|c|}{TopoCount \cite{abousamra2021localization}}  & \multicolumn{1}{c|}{61.2}          & \multicolumn{1}{c|}{104.6}  & \multicolumn{1}{c|}{7.8}          & \multicolumn{1}{c|}{13.7}          \\ \hline
\multicolumn{1}{|c|}{MBTTB-SCFB \cite{Sindagi_2019_ICCV8}}        & \multicolumn{1}{c|}{60.2}          & \multicolumn{1}{c|}{94.1}  & \multicolumn{1}{c|}{8.0}          & \multicolumn{1}{c|}{15.5}          \\ \hline


\multicolumn{1}{|c|}{S-DCNet \cite{Xiong_2019_ICCV}}        & \multicolumn{1}{c|}{58.3}          & \multicolumn{1}{c|}{95.0}  & 6.7          & 10.7          \\ \hline


\multicolumn{1}{|c|}{SSR-HEF(OURS)}                            & \textcolor{blue}{\textbf{55.0}}  &\textcolor{blue}{\textbf{88.3}}   & \textcolor{blue}{\textbf{6.1}}  &\textcolor{blue}{\textbf{9.5}}          \\ \hline
\end{tabular}
\end{center}
\end{table}

\begin{table}[]
\begin{center}
\centering
\caption{Estimation errors on the UCF\_CC\_50 and UCF-QNRF datasets.}
\label{Table:5}
\begin{tabular}{|c|c|c|c|c|}
\hline
\multicolumn{1}{|c|}{\multirow{2}{*}{Methods}}                                                 & \multicolumn{2}{c|}{UCF\_CC\_50}   & \multicolumn{2}{c|}{UCF-QNRF}  \\ \cline{2-5} 
                              & \multicolumn{1}{c|}{MAE}            & \multicolumn{1}{c|}{MSE}   & \multicolumn{1}{c|}{MAE}            & \multicolumn{1}{c|}{MSE}            \\ \hline
\hline

\multicolumn{1}{|c|}{Idrees et al. \cite{idrees2013multi} }                & \multicolumn{1}{c|}{419.5}          & \multicolumn{1}{c|}{541.6}  & \multicolumn{1}{c|}{315.0}          & \multicolumn{1}{c|}{508.0}     \\ \hline


\multicolumn{1}{|c|}{Wang et al. \cite{Wang_2019_CVPR}}                & \multicolumn{1}{c|}{373.4}          & \multicolumn{1}{c|}{528.8}  & \multicolumn{1}{c|}{-}          & \multicolumn{1}{c|}{-}         \\ \hline
\multicolumn{1}{|c|}{Wan et al. \cite{Wan_2019_CVPR}}        & \multicolumn{1}{c|}{355.0}          & \multicolumn{1}{c|}{560.2}  & \multicolumn{1}{c|}{-}          & \multicolumn{1}{c|}{-}         \\ \hline


\multicolumn{1}{|c|}{IG-CNN \cite{babu2018divide}}        & \multicolumn{1}{c|}{291.4}          & \multicolumn{1}{c|}{349.4}  & \multicolumn{1}{c|}{-}          & \multicolumn{1}{c|}{-}          \\ \hline   

\multicolumn{1}{|c|}{TEDnet \cite{Jiang_2019_CVPR}}        & \multicolumn{1}{c|}{249.4}          & \multicolumn{1}{c|}{354.5}  & \multicolumn{1}{c|}{113.0}          & \multicolumn{1}{c|}{188.0}          \\ \hline
\multicolumn{1}{|c|}{PACNN \cite{Shi_2019_CVPR}}        & \multicolumn{1}{c|}{241.7}          & \multicolumn{1}{c|}{320.7}   & \multicolumn{1}{c|}{-}          & \multicolumn{1}{c|}{-}         \\ \hline


\multicolumn{1}{|c|}{MBTTB-SCFB \cite{Sindagi_2019_ICCV8}}        & \multicolumn{1}{c|}{233.1}          & \multicolumn{1}{c|}{300.9}   & \multicolumn{1}{c|}{97.5}          & \multicolumn{1}{c|}{165.2}         \\ \hline



\multicolumn{1}{|c|}{S-DCNet \cite{Xiong_2019_ICCV}}        & \multicolumn{1}{c|}{204.2}          & \multicolumn{1}{c|}{301.3}   & \multicolumn{1}{c|}{104.4}          & \multicolumn{1}{c|}{176.1}        \\ \hline


\multicolumn{1}{|c|}{TopoCount \cite{abousamra2021localization}}  & \multicolumn{1}{c|}{184.1}          & \multicolumn{1}{c|}{258.3}  & \multicolumn{1}{c|}{89.0}          & \multicolumn{1}{c|}{159.0}          \\ \hline
\multicolumn{1}{|c|}{AMRNet \cite{2020Adaptive}}        & \multicolumn{1}{c|}{184.0}          & \multicolumn{1}{c|}{265.8}   & 86.6         & 152.2       \\ \hline


\multicolumn{1}{|c|}{SSR-HEF(OURS)}                            & \textcolor{blue}{\textbf{173.3}}  & \textcolor{blue}{\textbf{260.4}}   & \textcolor{blue}{\textbf{70.2}}  & \textcolor{blue}{\textbf{128.6}}         \\ \hline
\end{tabular}
\end{center}
\end{table}

\begin{table*}
\begin{center}
\centering
\caption{Estimation errors of different configurations.}

\label{Table:6}
\begin{tabular}{|c|c|c|c|c|c|c|c|c|c|c|}
\hline
\multicolumn{1}{|c|}{\multirow{2}{*}{Methods}}        & \multicolumn{2}{c|}{Part\_A}       & \multicolumn{2}{c|}{Part\_B}  & \multicolumn{2}{c|}{UCF\_CC\_50}  & \multicolumn{2}{c|}{UCF-QNRF} & \multicolumn{2}{c|}{WorldExpo’10}  \\  \cline{2-11}    
                                  & MAE           & MSE              & MAE           & MSE       & MAE           & MSE    & MAE         & MSE           & MAE           & MSE         \\  \hline
Without SSR strategy                 & 70.5                  & 108.6                   & 12.6                   & 20.5                   & 205.3             &  271.3        & 87.6              &157.2              & 8.9                 & 12.3         \\  \hline

Without  Feature Enhancement Layer        & 66.8                         & 108.0            & 9.4                       & 15.2          & 191.5                  & 288.3           & 79.8             & 140.5           & 7.6              & 9.5       \\  \hline

Without HEF algorithm        & 72.4             & 118.4                       & 15.1                & 25.1            & 226.2                 & 301.5           & 95.8             & 165.9           & 10.1            & 15.6       \\  \hline
SSR-HEF                      & \textcolor{blue}{\textbf{55.0}}    & \textcolor{blue}{\textbf{88.3}}           & \textcolor{blue}{\textbf{6.1}}           & \textcolor{blue}{\textbf{9.5}}     & \textcolor{blue}{\textbf{173.3}}        & \textcolor{blue}{\textbf{260.4}}  & \textcolor{blue}{\textbf{70.2}}    & \textcolor{blue}{\textbf{128.6}}  & \textcolor{blue}{\textbf{6.5}}    & \textcolor{blue}{\textbf{8.1}}\\ \hline

\end{tabular}
\end{center}

\end{table*}

\begin{figure}[t]
\begin{center}
    \includegraphics[width=1.0\linewidth]{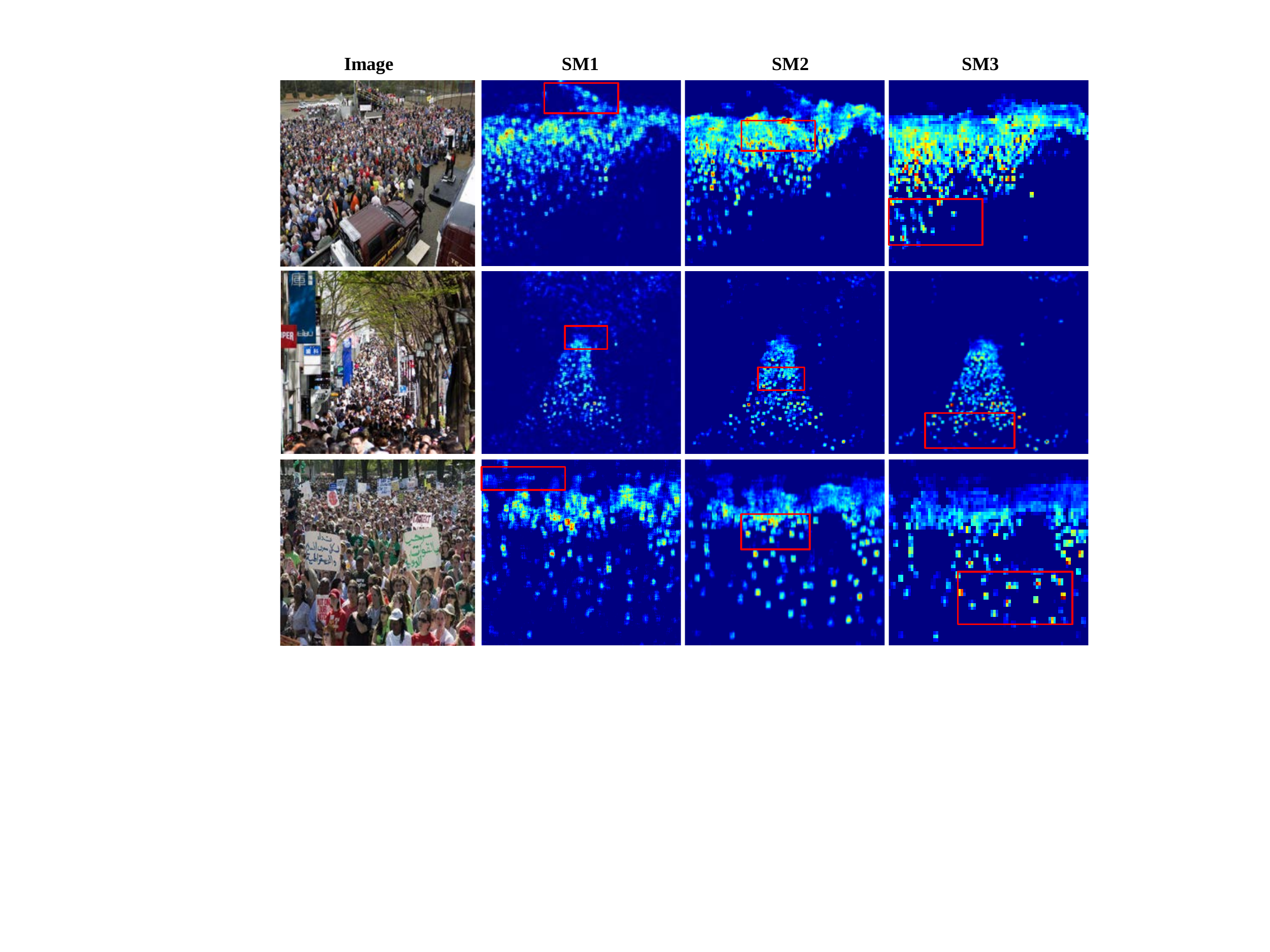}
\end{center}
   \caption{Some visualization results of the segmentation maps in different depth layers. Image column represents the test images. SM1, SM2, and SM3 Columns represent the segmentation maps of different depths.}
\label{FIG:7}
\end{figure}



\subsection{Evaluations and Comparisons}



\noindent{\bfseries ShanghaiTeach \cite{zhang2016single}.} This dataset consists of two parts: Part\_A and Part\_B. Part\_A contains a training set with 300 images and a test set with 182 images. All of them are collected from the Internet in a random way. Part\_B contains a training set with 400 images and a test set with 316 images. They are taken from the Shanghai streets. Since the density of Part\_A is higher than that of Part\_B, it is difficult to predict the crowd count accurately in this part.


Comparisons with other state-of-the-art methods are tabulated in Table \ref{Table:4}. LBP + RR are traditional feature extractors. Others are the CNN based methods. We can observe that the CNN based models outperform the traditional feature extractors (LBP+RR) significantly. In CNN based models, we can also find that the scale variation is extremely challenging and it limits the performance of proposed models. So many methods \cite{babu2018divide,Jiang_2019_CVPR,Shi_2019_CVPR,Sindagi_2019_ICCV8,Yang_2020_CVPR} are proposed to overcome it. Moreover, although the model (113M) \cite{Xiong_2019_ICCV} is 56 times larger than ours (2M), it is straightforward to see that the proposed SSR-HEF obtains the best result on both Part\_A and Part\_B.



{\bfseries WorldExpo’10 \cite{zhang2015cross}.} The WorldExpo’10 contains 3,980 frames sampled from 1,132 video sequences that are captured from 108 different scenes. The training set with 3,380 frames is produced from 103 different scenes, while the test set with 600 frames is generated from other 5 different scenes (Scene 1-Scene 5). The region of interest (ROI) has been given for each scene in this dataset. Following the experimental setting in \cite{zhang2015cross}, the provided perspective maps are also used to encode the ground-truth density map.

Following the recommendation of the author of this database \cite{zhang2015cross}, MAE is leveraged to evaluate the results of other state-of-the-art methods in Table \ref{Table:3}. It represents the accuracy of the method. In the same way, a clear review can be found that the CNN based methods outperform the traditional methods (LBP+RR). We can also observe that the proposed SSR-HEF leads the best performance in terms of Scene 1, 2, 4, 5, and average of five scenes, with $17\%$, $22\%$, $11\%$, $37\%$, and $17\%$ improvement over the second best method \cite{Shi_2019_CVPR}. Our proposed SSR-HEF is more robust, unlike others that will fail in a certain scene(e.g., Scene 4 in \cite{babu2018divide}). So our Average of 5 scenes is the best. The competitive result in Scene 3 is obtained. By reviewing the test set, we find that the ROI mask removes many hard examples in Scene 3. Hence, our strength has not been fully demonstrated.

{\bfseries JHU-CROWD \cite{Sindagi_2019_ICCV}.} The JHU-CROWD is an unconstrained dataset with 4,250 images that are collected under various weather-based degradations such as rain, snow, haze, etc. This dataset contains 1.11 million annotations at both image-level and head-level. Comparisons with other state-of-the-art methods are tabulated in Table \ref{Table:61}. We can observe that the proposed SSR-HEF outperforms other methods in all scenes, which shows the robustness of our proposed method.

{\bfseries UCF\_CC\_50 \cite{idrees2013multi}.} This dataset consists of 50 images randomly crawled from the Internet. Because of the challenging issues such as limited image, high density, and low resolution, the 5-fold cross validation is leveraged to evaluate the proposed methods.


We compare our method with other state-of-the-art methods in Table \ref{Table:5}. Traditional classical algorithms, such as SIFT, MRF, are used to extract crowd features in \cite{idrees2013multi}. Other methods on the ShanghaiTeach are still compared on this dataset. It can be also found that the deep learning outperforms the traditional algorithm \cite{idrees2013multi}. In addition, although \cite{2020Adaptive} uses the larger model (VGG-16) and is pretrained on the ImageNet dataset, we can observe that the proposed SSR-HEF achieves the best results.

{\bfseries UCF-QNRF \cite{Idrees_2018_ECCV}.} The UCF-QNRF is a large dataset with 1201 training images and 334 testing images. These high-resolution images contain a high density of crowd. There are 1,251,642 annotated heads in this dataset, with an average of 815 per image. The comparison results with recent state-of-the-art methods are given in Table \ref{Table:5}. In the same way, the performance of hand-crafted features \cite{idrees2013multi} is lost to deep learning methods. In deep learning methods, our SSR-HEF leads the best performance on both MAE and MSE, with the MAE/MSE 16.4/23.6 lower than that of second best performance \cite{2020Adaptive}.


\subsection{Ablation Study}
To further verify the effectiveness of our proposed method, we perform extensive ablation studies with analysis on these datasets, especially on the most popular ShanghaiTeach Part\_A \cite{zhang2016single}.

{\bfseries Benefits of Multi-Scale Semantic Refining Module:} To sufficiently overcome the scale variation in crowd scene, we propose a multi-scale semantic refining strategy. Multi-scale segmentation maps with semantic prior are first encoded, then they are mapped back to network to adaptively refine and extract multi-scale features. To demonstrate the effectiveness of this strategy, a set of comparative experiments is conducted. In comparative experiments, this strategy is removed from our model. $L_{segs}$ and corresponding parameters are also removed. Comparison results of our model with and without this strategy are elaborated in Table \ref{Table:6}. It can be observed that there is an evident utility in improving the performance by using this multi-scale semantic refining strategy. Since the scale variations of ShanghaiTeach Part\_A, UCF\_CC\_50, and UCF-QNRF are large, their improvements are more obvious.

\begin{table}
\begin{center}
\centering
\caption {Estimation errors about the classification task.} 

\label{Table:7}
\begin{tabular}{|c|c|c|c|c|c|}
\hline
categories                    & 0      & 5     & 10      & 15     & 20                 \\ \hline
\hline

MAE       & 68.2    & 63.4  & 61.8  &\textcolor{blue}{\textbf{55.0}}    & 65.0         \\ \hline
MSE       & 116.3   & 105.1  & 104.7  & \textcolor{blue}{\textbf{88.3}}   & 112.7      \\ \hline
\end{tabular}
\end{center}

\end{table}

\begin{figure*}[!t]
\begin{center}
\includegraphics[width=0.95\linewidth]{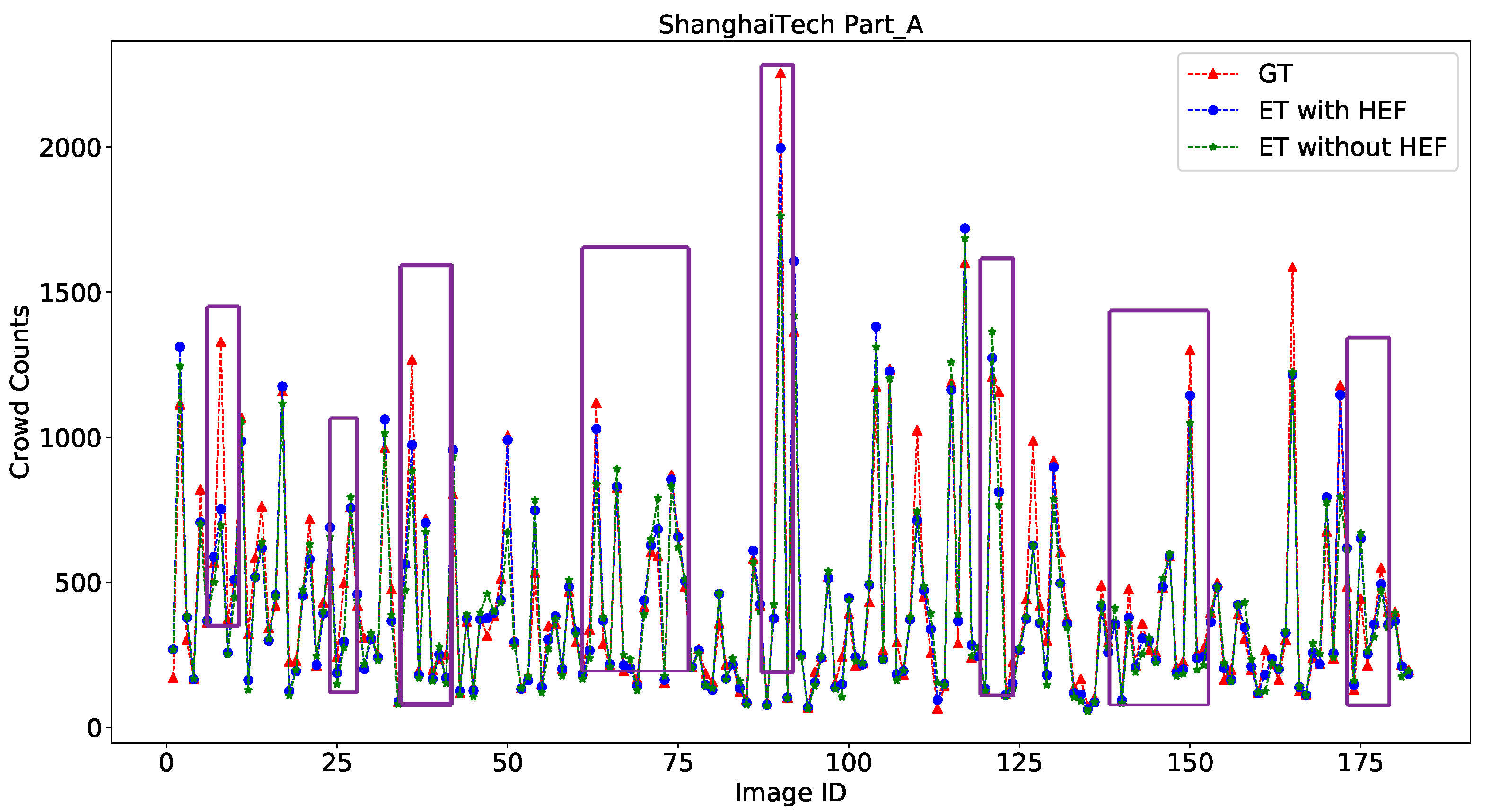}
   
\caption{The crowd count comparisons with and without the HEF. X-axis: the ID of images, Y-axis: the crowd counts of images.}
\label{FIG:8}
\end{center}
\end{figure*}

To further demonstrate the mechanism of this strategy, some predicted multi-scale segmentation maps in different depth layers on the ShanghaiTeach Part\_A are resized and visualized in Fig. \ref{FIG:7}. It can be observed that lower layers have learned some semantic concepts to roughly locate the crowd area. And layers of different depths play their respective advantages in feature extraction. As shown in red rectangles, small-scale crowds are captured in the SM1. Medium-scale crowds are highlighted in the SM2. Large-scale crowds are deeply depicted in the SM3, which is vital to refine the multi-scale features. Finding from this study and previous studies suggests that scale variations restrict the performance of crowd counting and effective extraction of multi-scale features is critical to overcoming the scale variation problem.

The auxiliary classification task digs into the level cue of density for performance improvement. To verify this effectiveness, comparative experiments with different classes are conducted on the ShanghaiTeach Part\_A. The results are given in Table \ref{Table:7}. '0' column represents the SSR-HEF without classification task. We can see that all results with the classification task are better than that without this task, which suggests that the classification task is helpful to improve the counting accuracy of the model. And the SSR-HEF with 15 classes leads the best result. We think that 15 classes are sufficient because of the limitations of model capacity.

{\bfseries Benefits of Feature Enhancement Layer:} To recognize the merits of this layer, comparative experiments with and without this layer are conducted. As shown in Table \ref{Table:6}, a clear review can be found that the SSR-HEF with this layer achieves higher accuracy on these datasets. We argue that crowd areas are enhanced in this layer, which is powerful to improve the accuracy of crowd counting.




\begin{figure}[t]
\setlength{\abovecaptionskip}{0.07cm}
\setlength{\belowcaptionskip}{-0.28cm}
\begin{center}
   \includegraphics[height=0.837\linewidth]{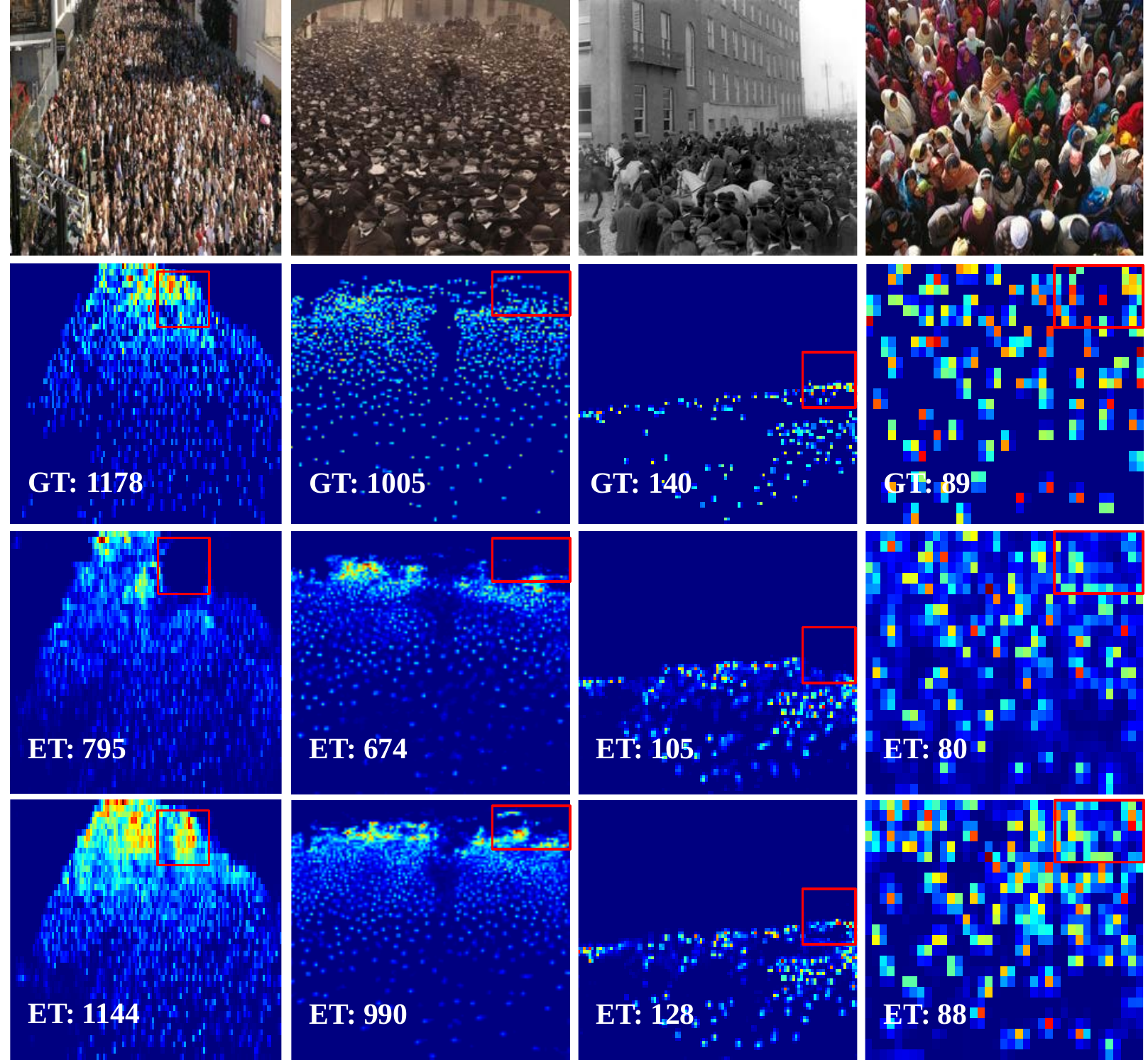}
\end{center}
   \caption{Some visualization results of the HEF algorithm. First Row: test image. Second Row: ground-truth density map. Third Row: the final estimated density map without hard example focusing algorithm. Fourth Row: the final estimated density map with hard example focusing algorithm. GT denotes the ground-truth crowd count. ET denotes the estimated crowd count.}
\label{FIG:9}
\end{figure}

{\bfseries Benefits of Hard Example Focusing Algorithm:} To set up a comprehensive evaluation of this algorithm, we perform a set of comparative experiments. In the comparative experiment, the HEF algorithm is removed and only the Euclidean distance is utilized to optimize the final density map. Comparison results are respectively elaborated in Table \ref{Table:6}. A clear review can be found that the performances are further enhanced by the proposed HEF algorithm on these datasets. 

To further visualize the effectiveness of this HEF algorithm on the ShanghaiTeach Part\_A, the final density maps respectively optimized by the Euclidean distance (third row) and our proposed HEF algorithm (fourth row) are illustrated in Fig. \ref{FIG:9}. As expected, the density (heat) of hard examples in red rectangles of the fourth row with the HEF is larger than that of the third row without the HEF. The density of hard examples in the red rectangle of the third row is mispredicted to be lower or even zero. The density maps of the fourth row and its ground-truth density map (second row) are more similar. And the crowd counting results of the fourth row are almost the same with the ground-truth counts, which further confirms that our proposed SSR-HEF focuses well on the hard example.

The predicted counting results of each image on the ShanghaiTeach Part\_A are shown in Fig. \ref{FIG:8}. We can observe that blue circles (estimated crowd counts with the HEF) are closer to red triangles (ground-truth crowd counts) than green pentagrams (estimated crowd counts without the HEF) in most images, which represents that the HEF is effective in reducing crowd counting errors. However, the performance of our algorithm is still limited by the number of labeled training samples. Inspired by \cite{9274525} where abundant unlabeled samples can be used for self-supervised learning, we will explore self-supervised methods to break the limitations to further improve the performance of our algorithm in future works.


\begin{table}[]
\begin{center}
\centering
\caption{Comparisons of the real time. Size represents the model size. Time represents the average time.}
\label{Table:8}
\begin{tabular}{|c|c|c|c|c|}
\hline
\multicolumn{1}{|c|}{Method}                             & \multicolumn{1}{c|}{Size}            & \multicolumn{1}{c|}{Time (s)}            \\ \hline
\hline

\multicolumn{1}{|c|}{AMRNet \cite{2020Adaptive}}        & \multicolumn{1}{c|}{226M}          & \multicolumn{1}{c|}{0.445}          \\ \hline
\multicolumn{1}{|c|}{Wang et al. \cite{Wang_2019_CVPR}}        & \multicolumn{1}{c|}{155M}          & \multicolumn{1}{c|}{0.417}          \\ \hline
\multicolumn{1}{|c|}{Wan et al. \cite{Wan_2019_CVPR}}        & \multicolumn{1}{c|}{66M}          & \multicolumn{1}{c|}{0.054}          \\ \hline
\multicolumn{1}{|c|}{S-DCNet \cite{Xiong_2019_ICCV} }        & \multicolumn{1}{c|}{113M}          & \multicolumn{1}{c|}{0.049}          \\ \hline
\multicolumn{1}{|c|}{TopoCount \cite{abousamra2021localization} }        & \multicolumn{1}{c|}{103M}          & \multicolumn{1}{c|}{0.009}          \\ \hline
\multicolumn{1}{|c|}{SSR-HEF}                            & \multicolumn{1}{c|}{\textcolor{blue}{\textbf{2M}}} & \multicolumn{1}{c|}{\textcolor{blue}{\textbf{0.003}}}          \\ \hline

\end{tabular}
\end{center}
\end{table}

\subsection{Evaluations of The Real Time:} To demonstrate the real-time effectiveness of our designed model, the comparative experiments of recent open source methods and ours are conducted with NVIDIA GTX 1080Ti and Intel Core i7. The test images of the ShanghaiTech Part\_A are resized to 512$\times$1024. Comparison results are tabulated in Table \ref{Table:8}. It can be observed that although three segmentation tasks introduce a few additional convolution operations, the model size of our designed model is still small (2M). And our model is about 3 times faster than \cite{abousamra2021localization} which is the second fast.

\section{Conclusion}
In this paper, we find the hard example problem in the regression task of crowd counting. The hard example focusing algorithm  is proposed to solve it. The lightweight SSR-HEF with this algorithm can rapidly focus on hard examples by automatically attenuating the weights of easy examples. To sufficiently address the scale variation problem, we first propose a novel multi-scale semantic refining strategy. Multi-scale segmentation maps with semantic prior are generated and mapped back to network to adaptively refine and extract multi-scale features. This strategy also makes lower layers of deep CNN capture the semantic concepts. Extensive experiments are conducted on six benchmark datasets to evaluate our method. Results imply that our method outperforms many state-of-the-art methods. Moreover, our designed model is smaller and faster. In the future, we will explore the self-supervised method to break the limitations of labeling cost and improve the performance of crowd counting.

\bibliographystyle{IEEEtran}
\bibliography{mulu_13_1}

\vspace{-10pt}

\begin{IEEEbiography}[{\includegraphics[width=1in,height=1.4in,clip,keepaspectratio]{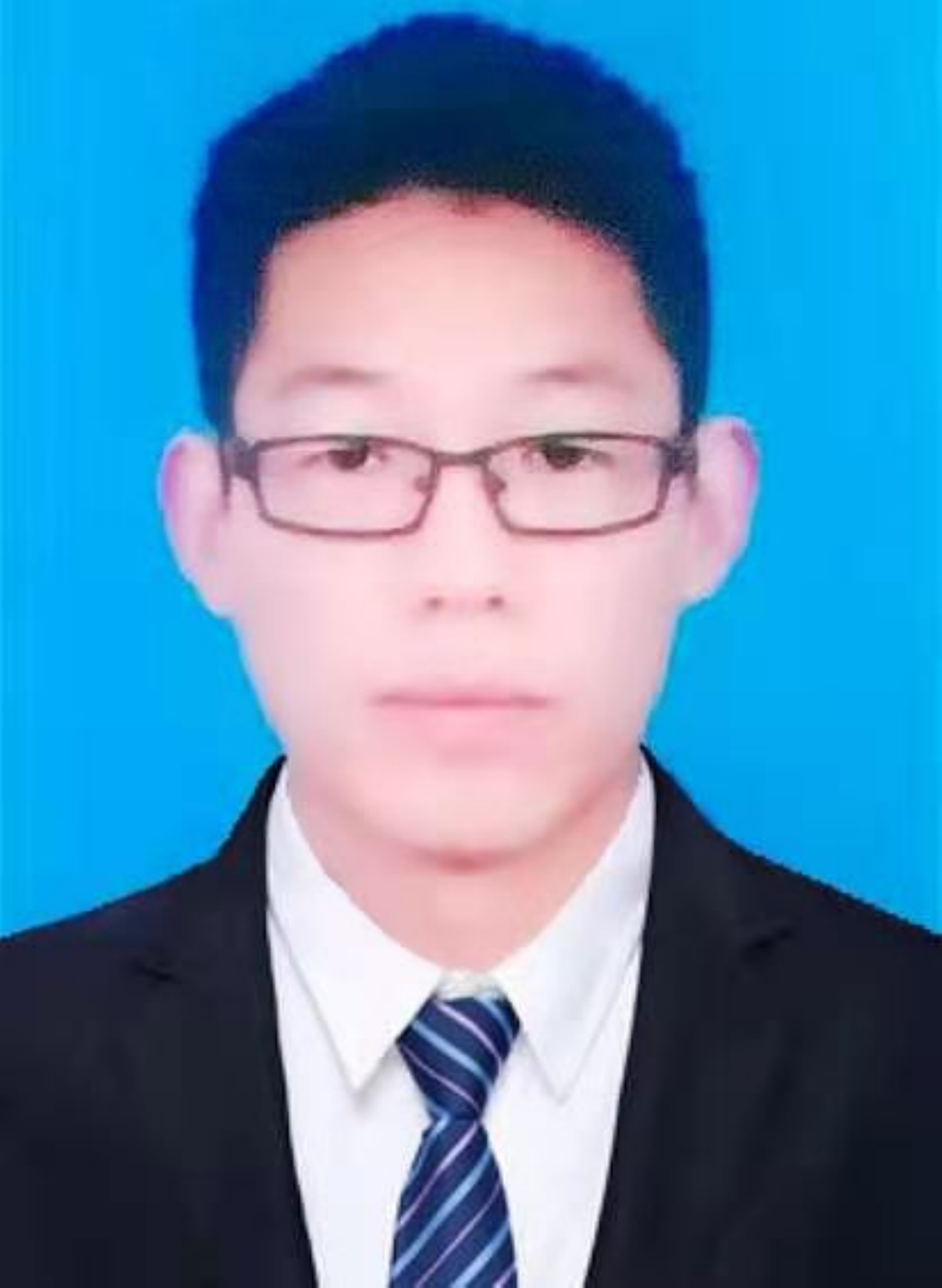}}]{Jiwei Chen} is currently pursuing the Ph.D. degree in the Hefei Institutes of Physical Science, University of Science and Technology of China. His research interests include computer vision, few-shot learning, semi-supervised learning and crowd counting.
\end{IEEEbiography}
\vspace{-35pt}


\begin{IEEEbiography}[{\includegraphics[width=1in,height=1.4in,clip,keepaspectratio]{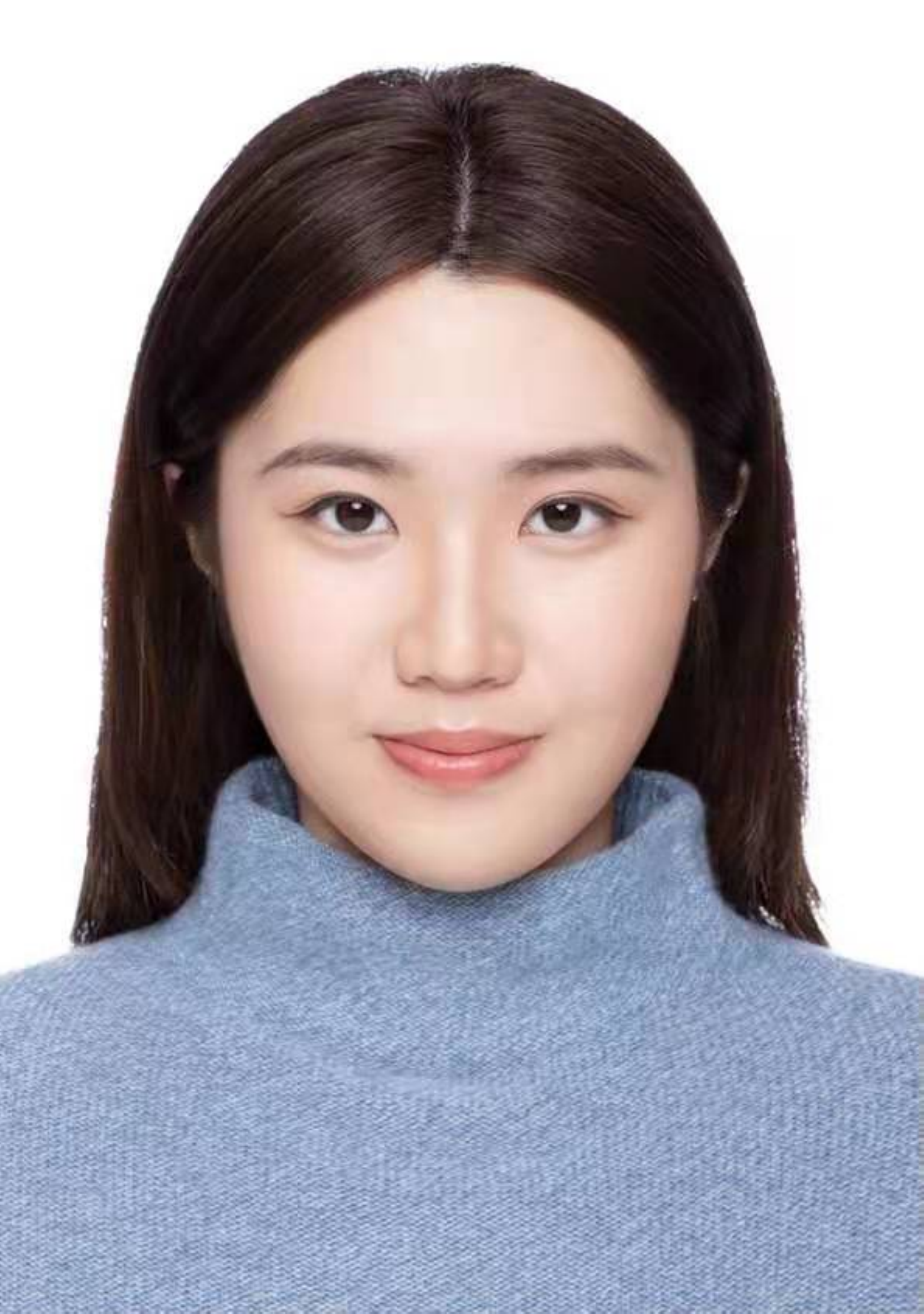}}]{Kewei Wang} is currently pursuing her M.Phil. degree at the School of Computer Science, Faculty of Engineering, the University of Sydney. Her research interests include computer vision, machine learning and crowd counting.
\end{IEEEbiography}

\vspace{-35pt}



\begin{IEEEbiography}[{\includegraphics[width=1in,height=1.4in,clip,keepaspectratio]{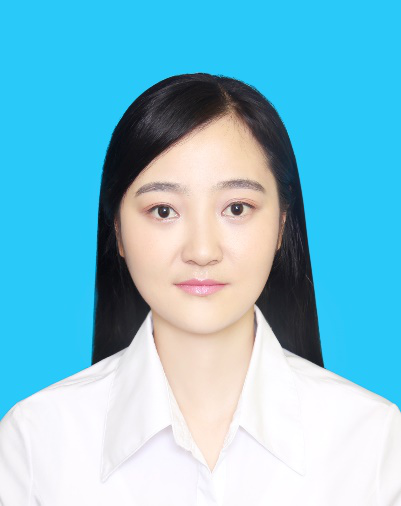}}]{Wen Su} received Ph.D. degree in control science and engineering from University of Science and Technology of China in 2018 and B.E. degree in engineering from Automation Department, University of Science and Technology of China in 2013, respectively. Now, she works in virtual reality laboratory in Zhejiang Sci-Tech University. At present her research interests are image segmentation and depth scene understanding based on deep learning.
\end{IEEEbiography}
\vspace{-35pt}

\begin{IEEEbiography}[{\includegraphics[width=1in,height=1.4in,clip,keepaspectratio]{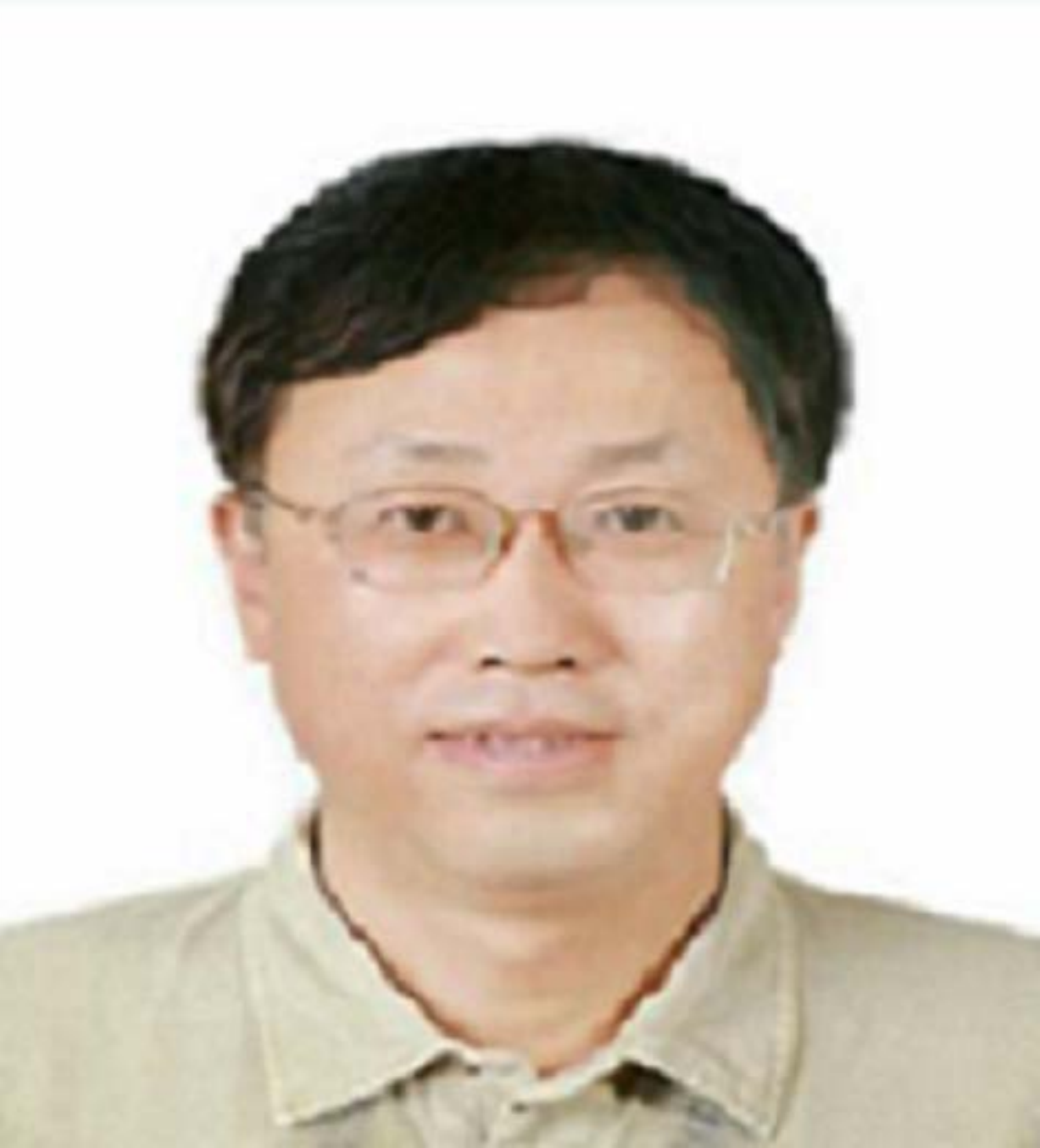}}]{Zengfu Wang} received the B.S. degree in electronic engineering from the University of Science and Technology of China in 1982 and the Ph.D. degree in control engineering from Osaka University, Japan, in 1992. He is currently a Professor with the Institute of Intelligent Machines, Chinese Academy of Sciences, and the Department of Automation, University of Science and Technology of China. He has published more than 300 journal articles and conference papers. His research interests include computer vision, human–computer interaction and intelligent robots. He received the Best Paper Award at the ACM International Conference on Multimedia 2009 and the IET Image Processing Premium Award 2017.
\end{IEEEbiography}

\end{document}